\documentclass[preprint,12pt]{elsarticle}

\usepackage{amssymb}
\usepackage{amsmath}
\usepackage{pdflscape}
\usepackage{booktabs}

\journal{Journal of Energy Storage}

\begin{document}

\begin{frontmatter}



\title{Decision-focused learning for optimal PV-Battery scheduling}


\author{Joris Depoortere} 
\author{Hussain Kazmi} 
\author{Johan Driesen} 

\affiliation{organization={ESAT-Electa KU Leuven},
            addressline={Kasteelpark Arenberg 10}, 
            city={Leuven},
            postcode={3001}, 
            country={Belgium}}

\begin{abstract}
The use of residential photovoltaics has increased dramatically in recent years. With battery systems becoming more affordable, the optimal operation of a photovoltaic-battery system can bring significant savings to households. Optimal control of these systems requires correct forecasts of the underlying parameters, such as photovoltaic power generation, to know how to schedule the battery. While forecasting models have become increasingly accurate due to algorithmic advances and data availability, accuracy is typically measured in generic metrics which might not align with the downstream application. This study proposes a decision-focused learning framework that integrates the optimization and prediction by training a Long Short-Term Memory photovoltaic energy forecaster on the downstream optimal scheduling of a battery system. The proposed methodology is compared against a standard two-phase approach. Across a 14-month evaluation period, the decision-focused method reduced average electricity costs across twenty buildings by 3.6\% when normalized against the performance bounds defined by a perfect forecast and a baseline of no optimization. Critically, this financial improvement was achieved despite the model exhibiting a root mean squared error of 19.9\%, significantly higher than the decoupled model's 8.2\%. Warm-starting the decision-focused model further improves the results, lowering the average cost by an approximate 8\% reduction, while also mitigating the negative impact on statistical accuracy (with a root mean squared error of 13.7\%). The findings are statistically significant at the 0.001 level following a Diebold and Mariano test across the twenty households and for each household individually. These results demonstrate that aligning forecast models with optimization goals is key for achieving cost advantages in PV-battery systems. Future research should aim to replicate these findings on other datasets, alternate forecasting models and alternate optimization algorithms.
\end{abstract}




\end{frontmatter}



\section{Introduction}
\label{sec1}
The integration of Renewable Energy Sources (RES) in the built environment has increased dramatically over the last decade. The EU enables this move towards distributed RES as it targets a 42.5\% share of renewables in our energy consumption under the Renewable Energy Directive \cite{ref1}. 
Within the built environment, residential households account for approximately 26\% of total energy consumption in the EU\footnote{Reported by Eurostat.}. Therefore, correct and sustainable integration of distributed RES in households will be critical to help the EU obtain its targets. Rooftop photovoltaics (PV) are one of the easiest to install RES for households. Bódis et al.\cite{ref2} estimate that the potential capacity of rooftop PV is approximately 680 TWh per year in the EU, covering approximately 24\% of our total consumption (in 2019). 


However, PV systems are intermittent in nature, and on top of their intermittent nature they are negatively correlated with home energy consumption, as most energy in residences is consumed in the morning and the evening. These two factors pose operational challenges in the effective integration of PV systems into the distribution grid. Battery Energy Storage Systems (BESS) can help alleviate these challenges, allowing us to control the intermittent energy by storing it for when it is needed, typically in the evening. But BESS are still expensive investments for residential consumers, and in order to efficiently make use of a BESS, optimal operation of the system is required, discharging the BESS mainly when energy prices are high and charging when prices are low. 

Dynamic price contracts help with cost efficient operation of BESS. These dynamic price contracts reflect wholesale electricity prices and are communicated to clients a day in advance. This way, the prices are different for each hour of the day and better aligned with the availability of electricity and its consumption. In Belgium it will become mandatory from 2027 onward for energy suppliers to provide customers with the option of such a dynamic price contract\footnote{see https://vrtnws.be/p.nwq0QwAQx.}. Optimally running a PV-Battery system with these dynamic contracts is invaluable to make the battery a worthwhile investment. This is a complex task as several contextual parameters are required to properly schedule the battery. While prices are known a day in advance, the electricity demand of the household as well as the photovoltaic energy generated are uncertain and require accurate forecasting to generate a correct schedule. In this study we propose to integrate the task, forecasting photovoltaic energy based on the cost savings from optimizing a residential battery.

 In the next chapter we will elaborate on recent research in these fields. First, recent literature on residential PV forecasting is discussed, followed by research on PV-Battery optimization. Finally, we look at the intersection of forecasting and optimization, called decision-focused learning, what the developments are in this field, and where in the field of electrical energy it has been applied. The remainder of this paper will go over the data that was used to conduct the experiments, the methodology of our research, i.e. the optimal scheduling problem, how we apply decision-focused learning and how it differs from a regular two-step approach, our main results, which showcase the RMSE of the models and, more importantly, the downstream loss. And finally, we end with a discussion and conclusion in which we also propose several future directions.

\section{Literature review}

Research in the field of optimal BESS scheduling can be split into the two stages that compose the scheduling problem: First, obtaining all the required contextual parameters for solving said problem, which includes several stochastic parameters requiring forecasting, for example the PV output and the household energy demand. And second, the execution of the optimization problem to obtain an optimal battery (dis)charging schedule, requiring an objective to optimize, correct handling of all the necessary constraints, and execution using a suitable optimization algorithm. 

\subsection{PV Forecasting}

Research that looks at forecasting PV (or load) typically does so in a stand-alone manner, without considering the objective at hand. A lot of modern research does so in a data-driven manner, with the increased popularity of machine learned (ML) models, such as boosting and bagging methods or neural networks (see for example: \cite{ref3}, \cite{ref4}, \cite{ref5}, \cite{ref6}, \cite{ref7}, \cite{ref8}, \cite{ref9}). 

PV forecasting is such a prominent field that several review papers are published on a yearly basis on the most recent developments. In \cite{ref10} Barhmi et al. discuss the role of artificial intelligence on solar forecasting. They mainly focus on which models are frequently used, depending on the time horizon (from ultrashort 1 minute forecasts to longer forecasts of up to 24 hours). They go on to discuss that most modern research relies on recurrent neural networks and the Long Short-Term Memory (LSTM) model, or a variant thereof. For example, Lim et al.\cite{ref11} have found a lot of success combining an LSTM with a Convolutional Neural Network (CNN). Barhmi et al. go on to discuss error metrics prevalently used to evaluate PV forecasting. Interestingly, all of the performance evaluation metrics mentioned are statistical metrics, with the most prevalent ones being the (Root) Mean Squared Error (RMSE), the Mean Absolute Error (MAE) and the Forecast Skill (which basically takes a statistical metric and evaluates it relative to a benchmark model). Lim et al. also evaluated the performance of their CNN-LSTM model based on the MAE, RMSE, MAPE (the Mean Absolute Percentage Error) and the R². 

A typical pipeline for the development of such a data-driven forecast model requires the use of data from the target field to train and validate an ML model on, often augmented with relevant features in the target space (for example values for cloudiness when PV is forecast). This model is then trained based on one of the generic statistical metrics we mentioned before, such as the RMSE \footnote{See Yang et al. \cite{ref12} for a further discussion on statistical error metrics in solar forecasting.}. Using this statistical metric, one obtains a single value that shows how strongly the forecast values, $\hat{y}_i$, deviate from the true values, $y_i$. Ideally, it does not matter that we use the MSE as error metric to train our model, as an MSE of zero will give us the perfect forecast. Similarly, if the error metric is the Mean Absolute Error (MAE), a value of zero again results in a perfect, and therefore identical, forecast to the RMSE. However, due to the inherent stochastic nature of forecasts, there will always be a non-zero error, and therefore the forecast is susceptible to biases induced by the choice of error metric. 

\subsection{Two-stage optimization}

Studies that look at the implementation of algorithms and the downstream objective being pursued are plentiful, but accurate forecasts (or forecasts representative for state of the art methods) are usually not the focal point in these studies. In \cite{ref13} Nottrott et al. optimize a PV-Battery (PV-B) system for peak management and cost minimization. They use a simple moving average for their PV forecast and for the load forecast they incorporated a random noise component of 5\% for their predictions, which is far too low for individual residential load forecasts. In \cite{ref14}, Zou et al. use an average error of 15\% for residential load, based on the idea that state-of-the-art models might be able to achieve this kind of error. Zhang and Tang obtain their solar PV forecasts from third party providers without much elaboration regarding the quality of these forecasts \cite{ref15}. Song et al. \cite{ref16} show in their study that predictions are made using a neural network in their home energy management system, but again nothing more is said about these predictions and their precision. In \cite{ref17} Luo et al. claim to use a forecast-based operating strategy for a PV-B system, but even so, they also use external forecasts for PV and a simple persistence model for load, whereby the load of the previous day is the forecast of the next. Finally, in \cite{ref18} Liang et al. evaluate the performance of a two-stage scheduling framework, first evaluating the economic performance of the schedule and then changing the downstream objective in the second stage to account for the impact of outages. They used a mixed integer linear program for their optimization and model the outage probability with a Markov Chain. But again, the role of the forecasting model is neglected, as PV, load and weather forecasts are inputs to the optimization, but no detail is provided on the quality of these forecasts, even though the study sees resilience as a key objective which is highly impacted by the accuracy of the forecasts. 

The above clearly shows the gap in research on optimal PV-B system scheduling. Either the focus lies on the scheduling itself, and forecasts used as inputs for the optimization are too simple or unfairly accurate (by adding a small noise to actual data or by using simulated data); or the focus lies solely on the forecasts; and these forecasts are decoupled from the optimization problem at hand by evaluating them on generic metrics without considering the use case further down the line, introducing biases based on the chosen metric, which might not align well with the downstream use case. 

\subsection{Decision-focused learning}

Recent research in machine learning has begun to bridge this gap between forecasting and optimization. This area is still in its infancy, and no standard nomenclature has yet emerged. When the focus is on optimization, the term end-to-end learning is often used, whereas in forecasting using deep learned neural networks, decision-focused learning has gained some traction, although several alternative terms are also in circulation.\footnote{Other commonly used terms include value-oriented forecasting and predict-then-optimize.} We will use the term decision-focused learning throughout the remainder of this article.

Decision-focused learning (or DFL) ultimately boils down to the integration of predicting uncertain parameters and optimally executing a problem that requires these parameters as input. Training the prediction model is done by evaluating the output of the downstream optimization problem, not the output of the model itself. In the end, a perfect forecast will bring with it the lowest costs (or highest reward) for the optimization, as any deviation from the actual parameters will cause the optimization to be performed with 'faulty' information, thus obtaining suboptimal solutions. But forecasts are never perfect, and the goal behind DFL is the prioritization of a better optimization objective over statistical accuracy, which is what we are ultimately interested in. 

Kotary et al.\cite{ref19} discuss this setting (they call it Predict-and-Optimize) in their survey on end-to-end learning. The typical way to train in a DFL setting is by introducing an external optimization solver into the training of the model. Agrawal et al.\cite{ref20} executed this in their work on differentiable optimization layers, where they extract gradients for backpropagation by implicit differentiation of the Karush–Kuhn–Tucker (KKT) conditions of a quadratic program. The disadvantage of this method is the inclusion of an optimization solver in each backward pass, which comes with a high computational cost. Pascal Van Hentenryck discusses this topic in \cite{ref21} in which he proposes the use of optimization proxies with dedicated 'repair' layers. These neural networks go straight from the input space to a solution for the optimization problem itself, which goes through a repair layer to return the solution back to the feasible space given key constraints of the optimization problem. This bypasses the need for implicit differentiation of the optimization problem. This optimization proxy method is advantageous for complex optimization problems for which implicit differentiation would be too much of a computational burden, which is likely overkill for a simple PV-B system. 

Interestingly, Bergmeir et al. \cite{ref22} held a competition on forecasting and optimal control of a PV-B system on a university campus. The competition included forecasting building load and PV production as well as optimizing the operation of the battery system. Notably, none of the participants used a decision-focused framework in the forecasting section of the competition. All teams made separate models based on features such as the weather and lagged target values. 

However, some work has been done in this modern research field, within the energy space. Paredes et al. \cite{ref23} used DFL to predict automatic Frequency Restoration Reserve (aFRR) capacity prices on the Belgian market to facilitate the participation of a large BESS on the reserve markets. They train their LSTM-like forecasting model on a hybrid loss which accounts for both the prediction error and the downstream profit loss. The BESS was a stand-alone system and no uncertain parameters other than the aFRR are considered for the contextual optimization problem. In their work they claim a profit improvement of 9.47\% using the DFL method. Depoortere et al. \cite{ref24} present an initial study of DFL in a PV-B scenario. Their work demonstrated the potential of the approach but relied on several simplifying assumptions: household load was assumed to be known in advance and no weather data were included in the PV forecast. Both of these can significantly impact the results. Additionally, the analysis did not examine in detail where and how the DFL forecasts provided value compared to generic forecasts. Table \ref{tab:lit_comparison} gives a comparison of the relative improvements these studies obtained compared to two-stage forecasting and optimization within their fields, as well as some additional studies in other related energy fields (\cite{ref25}, \cite{ref26}, \cite{ref27}, \cite{ref28}, \cite{ref29}).

\begin{landscape}
\begin{table}[htbp]
\centering
\caption{Comparison of DFL application areas, methods and performance gains in literature}
\label{tab:lit_comparison}
\begin{tabular}{|
    p{3cm} |
    p{3cm} |
    p{3cm} |
    p{2.5cm} |
    p{2.5cm} |
    c |
}
\hline
Study & Forecast & Application & Method & Baseline & Cost / Profit \\
 &parameter & & &model&improvement \\
\hline
Paredes et al. (2025) & aFRR prices & Balancing market participation & Differentiable optimization & LSTM & 9.47\% \\ \hline
Zhang et al. (2025) & Wind power generation & Operating costs & Differentiable optimization & Neural network & 9.5\% \\ \hline
Depoortere et al. (2025) & PV power generation & PV-Battery scheduling & Differentiable optimization & LSTM & c.\ 5\% \\ \hline
Dupont et al. (2024) & Day-ahead (DA) Electricity prices & DA scheduling & SPO+ & LSTM & c.\ 1\% \\ \hline
Wahdani et al. (2023) & Wind power generation & System dispatch & Differentiable optimization & GRU & c.\ 10\% \\ \hline
Han et al. (2021) & Load & Economic dispatch & Task loss & Neural network & 0.24-0.27\% \\ \hline
Stratigakos et al. (2021) & Imbalance prices & Balancing market participation & SPO & Random Forest & 0.58\% \\ \hline
\end{tabular}
\end{table}
\end{landscape}

\subsection{Contributions}

Given the limitations that we see in current research on residential PV-B scheduling, and how it is always done in a two-phase approach, whereby the phase not under consideration (forecasting when the focus is on optimization and vice versa) is typically ignored or executed using simple methodologies, we propose an integrated pipeline of forecasting and optimal PV-B scheduling, following the DFL paradigm using implicit differentiation. In this study, we focus on forecasting PV, using a neural network which is trained and evaluated based on the downstream task of optimally charging and discharging a BESS to minimize the daily cost of electricity. We apply this framework to 20 different buildings over a span of multiple years to obtain results on a statistically significant sample size to further substantiate our findings. 

Our contributions in this space are as follows:
\begin{enumerate}
    \item We perform a deep-dive study of the impact of DFL on the value created in a PV-Battery system. We provide a detailed comparative study, evaluating the performance of the forecast based on the RMSE, a generic statistical metric, and based on the downstream cost savings in a PV-B system.
    \item Furthermore, we apply a combined methodology of warm-starting a DFL model by first training it on the RMSE and fine-tuning it in a decision-focused manner afterwards. We do this to avoid cold-start issues which may arise from decision-focused learning. These two form our basis experiments discussed in section \ref{base}.
    \item Finally, we go beyond assessing the baseline effectiveness of decision-focused learning by looking at scenarios where other parameters of the optimization problem become increasingly uncertain, as these parameters directly influence the training of the forecaster by being included in the online optimization layer. In our case, the household load is a prime example of such a parameter. This analysis is done in section \ref{load}.
\end{enumerate}

\section{Data}

In this section we discuss the data used for our experiments. First, we go over the actual buildings for which we have 3 years of smart-meter data from 2018 up until 2021 containing our load and PV generation data. Next, we elaborate on the data which was generated for the purpose of this study. The houses do not have a battery system, so one was simulated for them. We also created a price profile to reflect a dynamic price contract for the households.  

\subsection{The buildings}

\begin{figure}[!t]
\centering
\includegraphics[width=\textwidth]{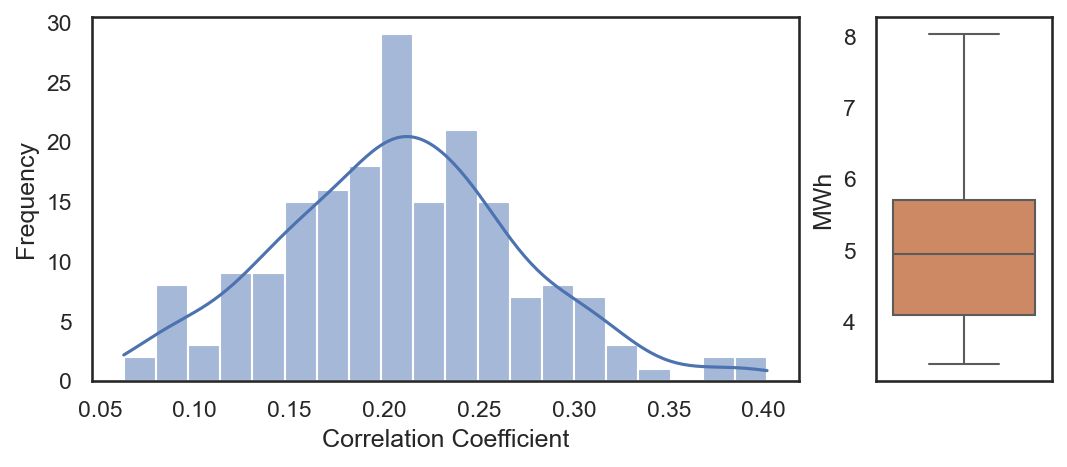}
\caption{Summary statistics for the 20 buildings. Figure 1(a) plots a histogram of the correlation coefficients between each of the buildings and Figure 1(b) depicts a boxplot of the maximum yearly load for the 20 buildings over the 3 year period.}
\label{fig1}
\end{figure}

The data used in our study comes from a neighborhood of 20 buildings located in Utrecht, in the Netherlands. The houses are part of a net-zero energy project, undergoing a deep retrofit in 2017, which makes them very similar from a design perspective. All houses have an identical PV setup, with 6 kWp PV panels. The lowest correlation between the time series of PV power output is 0.96. The load profiles, on the other hand, are very different. Figure \ref{fig1}(a) plots a histogram of the correlation matrix of the load time series for all the houses. Correlations range between 0.06 and 0.4. The size of the load profiles is also very different across the buildings. Figure \ref{fig1}(b) shows the spread in total annual consumption among households in MWh, ranging between approximately 3.5 to 8 MWh. 

\subsection{Generated data}

\begin{figure}[!t]
\centering
\includegraphics[width=\textwidth]{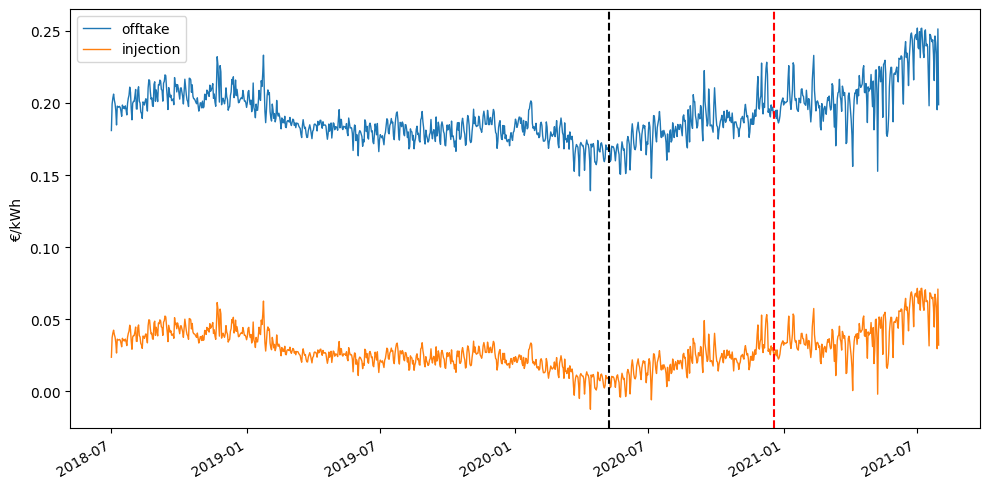}
\caption{Generated price profile across the households. Both off-take (blue) and injection (orange). The black vertical line indicates the 60/40 train and test set split while the red vertical line indicates what an 80/20 split would look like.}
\label{fig2}
\end{figure}

To implement optimal PV-B scheduling, several other parameters are needed. We need to know the hourly electricity prices, and we require meta-data regarding the battery being used. For the electricity prices, we generate a time-series based on an example formula used for dynamic contracts\footnote{obtained from https://www.mijnenergie.be/blog/dynamische-tarieven-2023-hoeveel-betaal-je-met-een-dynamisch-tarief in April 2025}. Additionally, for off-take prices we add base taxes and Opslag duurzame energie- en klimaattransitie (ODE) taxes\footnote{Following the taxes given at https://www.belastingdienst.nl}. Figure \ref{fig2} plots the off-take and injection prices for our proposed contract over the time period. It is worth noting the rise of the prices from 2021 onwards, which can have an impact on training, which is why we use a conservative train/test split of 60/40. Typically, studies split data at a ratio of 80/20. Both our 60/40 and the 80/20 split are plotted on the graph. As shown, an 80/20 split would make it so the price regime in the test data is very different from the train data. Additionally, an 80/20 split would only include half a year to evaluate on while a 60/40 split would allow us to evaluate the model on a full year (from May'20 until Jun'21).

We have no (meta-) data available of batteries being present in the buildings, so we simulate our own set-up. For the capacity of the battery, we use Weniger et al.\cite{ref30}, who propose a ratio of 1.1 kWh/MWh of battery size over yearly load. For the c-rate of the battery we use 2.7, meaning the battery cannot (dis)charge more than 0.37 times the capacity of the battery at any given hour. This C-rate is based on a Tesla Powerwall\cite{ref31}, a commercially available BESS. These two parameters are enough to define our battery; we do not delve further into efficiency curves or battery losses for this study as it is not our focus.


Finally, we make use of some input features that help with forecasting PV generation. Irradiance, cloudiness, and several other weather parameters are commonly used to forecast PV generation. We limit ourselves to direct normal radiation (DNI), one of the key weather parameters in predicting PV. We obtain this data from open-meteo\cite{ref32} in the form of historical actuals, but use them as forecasts. We do this because historical forecasts are notoriously difficult to obtain due to their dimensionality (not only do you need the values for specific hours, you also want a specific lead time). To avoid data leakage by providing our model with actual historical values, we add several noise factors to these actuals to make them into more realistic forecasts. The way in which noise is added to the actuals is a compound error process as given by Eq. \ref{eq:mult_noise} to Eq. \ref{eq:smoothing}.

\begin{align}
\tilde{y}_t &= y_t \left( 1 + \beta z_t \right),
\quad z_t \sim \mathrm{SkewNormal}(\alpha),
\quad \forall t
\label{eq:mult_noise}
\\
\tilde{y}_t &\in \left[ 0, \max(y) \right],
\quad \forall t
\label{eq:clipping}
\\
t' &= t + \Delta_t,
\quad 
\Delta_t =
\begin{cases}
U(-k,k), & \text{w.p. } p  \\
0, & \text{otherwise}
\end{cases},
\quad \forall t
\label{eq:time_shift}
\\
\hat{y}_t &= \sum_{\tau} \tilde{y}_{\tau}
\frac{1}{\sqrt{2\pi}\sigma_s}
\exp\!\left( -\frac{(t-\tau)^2}{2\sigma_s^2} \right),
\quad \forall t
\label{eq:smoothing}
\end{align}

\begin{figure}[!t]
\centering
\includegraphics[width=4in]{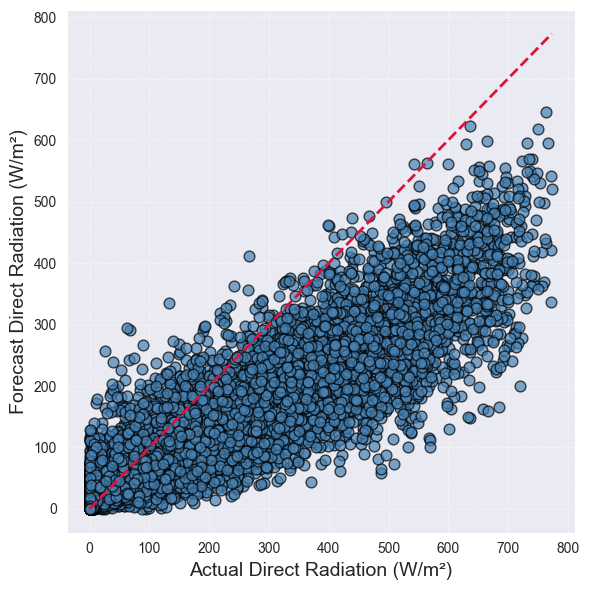}
\caption{Generated DNI forecasts based on actual, historical DNI data. The DNI data is made noisy on purpose to avoid data leakage.}
\label{fig3}
\end{figure}

 First a skewed noise is added to the values. Irradiance forecasts from data-driven models are typically skewed negatively because irradiance models trained on the MSE tend to underpredict to avoid large errors, a bias inherent in the MSE. The $\alpha$ in Eq. \ref{eq:mult_noise} indicates the direction of this skewness and $\beta$ is used to determine the scale of the error. Next the values are clipped (Eq. \ref{eq:clipping}); this way they stay within physical bounds after adding noise. As a third step, random timing shifts are introduced to simulate the model predicting peaks too soon or too late following a discrete uniform distribution (Eq. \ref{eq:time_shift}). Finally, a Gaussian filter is applied to introduce realistic autocorrelation in the error (Eq. \ref{eq:smoothing}). For the DNI forecasts we used a scaling factor of 20\%, an $\alpha$ of -1, a probability of time shifting of 2\% with a maximum shift of 5 hours and a Gaussian smoothing factor of 1. Figure \ref{fig3} shows a scatterplot of the actual DNI versus our generated forecasts. The normalised RMSE of the forecast data is 0.098, which is on the higher end of accuracy, given modern research. But accurate DNI forecasts benefit both models trained on RMSE as well as models trained in a DFL framework, so this should have limited impact on the results. 

\begin{figure*}[!t]
\centering
\includegraphics[width=\textwidth]{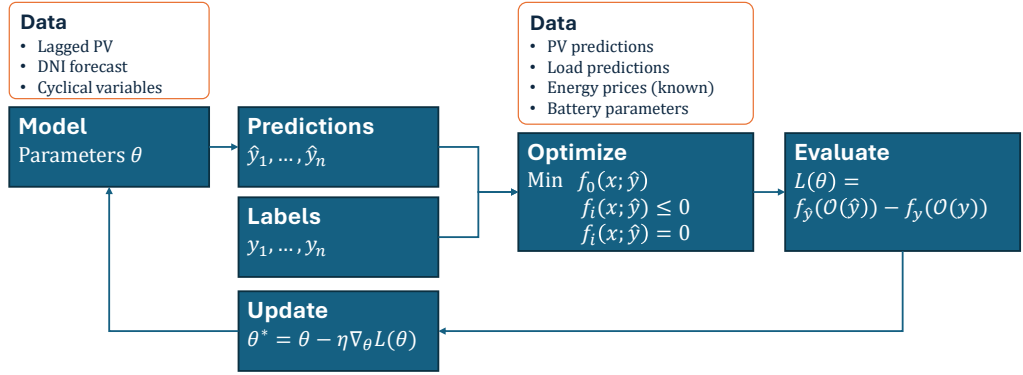}
\caption{Framework for training a DFL forecaster with the data needs for the forecasting model and the optimization problem. The Optimize step is done online during training and produces variables to obtain the cost given the forecast which gets evaluated against the cost given the real PV. This difference is used to update the gradients of the forecast model.}
\label{fig4}
\end{figure*}

This concludes the data needed to execute the DFL framework. Figure \ref{fig4} gives a summary of the data and where it is needed. We revisit the flow in this figure when we discuss the training of a DFL forecaster in Section \ref{DFL}.

\section{Method}

In this section, we discuss the methodology applied in this study. First, we give a general overview of the workflow. Then, we go over the convex optimization problem which forms the core of our interest and the basis of why we are forecasting to begin with. We discuss this first because the optimization problem is also included as a step within the training procedure of our forecaster. Next, we elaborate on the forecasting models used to predict PV generation. Finally, we discuss the evaluation metrics used to assess the performance of the forecasting models. 

\subsection{Workflow}

We are interested in optimizing the schedule of our PV-B system. More specifically, this means that we want to optimally charge and discharge the battery so as to pay the least amount of money for consumed energy. We do this in a daily manner, meaning we want to minimize the daily energy cost. In order to do so, we require a forecaster that gives us predictions for the next 24 hours, starting at midnight. We limit ourselves to this case for our main experiment and thus do not update the schedule during the day. During execution we therefore have the following flow:
\begin{itemize}
    \item The forecasting model is given the actual PV generation for the last 24 hours, a forecast of DNI for the next 24 hours, and cyclical variables indicating the hours of the day.
    \item The forecaster then provides us with a 24 hour forecast of PV generation for the next day.
    \item This forecast is used as an input for the optimization solver, together with energy price data and load forecast data for the next 24 hours. 
    \item The solver runs and we are given a 24 hour schedule for charging and discharging the battery.
\end{itemize}

To execute this workflow we require two key ingredients: an optimization problem for the PV-B system and a forecasting model for the PV generation. These two are discussed in the next subsections. 

\subsection{Convex optimization}

The library we use for differentiable optimization as a neural network layer, cvxpylayers\cite{ref20}, requires the use of convex optimization to work. Convex optimization is of the form

\[
\begin{aligned}
\min_{x} \quad & f_0(x; \hat{y}) \\
\text{s.t.} \quad & f_i(x; \hat{y}) \leq 0, \\
& f_i(x; \hat{y}) = 0
\end{aligned}
\]

Where the objective function and the inequality constraint functions must be convex and the equality constraint functions must be affine. Convex optimization does limit us in the complexity of the PV-B optimization problem we are considering. For example, we cannot force our battery to only charge or discharge every hour, as that requires the introduction of a boolean value, which would transform the problem to a mixed integer linear program. However, we do have some ways available to us to work around this issue. Our PV-B optimization problem is as follows:

\paragraph{Objective Function}
\begin{align}
\min_{\mathbf{P_{im}}, \mathbf{P_{ex}}} \sum_{t=1}^T (P_{im,t} \cdot Pr_{im,t} - P_{ex,t} \cdot Pr_{ex,t})
\label{obj_func}
\end{align}

\paragraph{subject to}
\begin{align}
\mathbf{P_{pv}} + \mathbf{P_{im}} + \mathbf{P_{dis}} = \mathbf{P_{ex}} + \mathbf{L} + \mathbf{P_{ch}}, \quad \forall t
\label{balance}
\end{align}
\begin{align}
\mathbf{P_{ex},P_{im},P_{ch},P_{dis}} \geq 0 \quad \forall t
\label{nonneg}
\end{align}
\begin{align}
\mathbf{P_{ch}} \leq \mathbf{C_{bat}/2.7 \cdot mode}, \quad \forall t
\label{ch_lim}
\end{align}
\begin{align}
\mathbf{P_{dis}} \leq \mathbf{C_{bat}/2.7 \cdot (1-mode)}, \quad \forall t
\label{dis_lim}
\end{align}
\begin{align}
0\leq \mathbf{mode_{t}} \leq 1, \quad \forall t
\label{SOC}
\end{align}
\begin{align}
0.2 \cdot \mathbf{C_{bat}} \leq \mathbf{SoC_{bat,t}} \leq 0.8 \cdot \mathbf{C_{bat}}, \quad \forall t
\label{SOC}
\end{align}
\begin{align}
\mathbf{SoC_{bat,t}} = 
\begin{cases}
\mathbf{0.5 \cdot C_{bat}}, & \text{if } t = 0, T \\
\begin{array}{@{}l@{}}
\mathbf{SoC_{bat,t-1}} + \mathbf{P_{ch,t-1}} - \mathbf{P_{dis,t-1}}
\end{array}, & \text{if } 0 < t < T
\end{cases}
\label{SOC_time}
\end{align}

The downstream objective (Eq. \ref{obj_func}) is the minimization of the daily cost, which is obtained by summing the hourly cost. The cost comes down to the imported energy ($P_{im}$) multiplied by the cost of imported energy (($Pr_{im}$) minus the exported energy ($P_{ex}$) multiplied by the profit from exporting energy to the grid ($Pr_{ex}$). 

The optimization problem is constrained by several factors. First of all, there is the energy balance constraint, Eq. (\ref{balance}) which states that the uses of energy (the load ($L$), exported energy and the battery charging ($P_{ch}$)) should equal the supply (generated PV ($P_{pv}$), imported energy and battery discharging ($P_{dis}$)). Next, Eq. (\ref{nonneg}) deals with the non-negativity of power exported, imported, charged and discharged (the PV and load are given parameters so a check for non-negativity is not needed). Eq. (\ref{ch_lim}) and Eq. (\ref{dis_lim}) limit how much power can be charged and discharged from the battery at any given time step. The factor 2.7 is taken from \cite{ref31}, as mentioned before. Using a 'mode' variable allows us to limit simultaneous charging and discharging every hour, without having to introduce dummy variables, maintaining convexity. The capacity, $C_{bat}$, of the battery has also been discussed in the previous section. Eq. (\ref{SOC}) limits the state of charge of the battery ($SoC_{bat}$) to be between 20\% and 80\% of the total capacity, and Eq. (\ref{SOC_time}) is used for the calculation of the SoC of the battery over time. The initial SoC is chosen to be 50\% of the capacity of the battery. We also set the battery back to the initial state at the end of the 24-hour period to avoid that the solver will just drain the battery completely during the day.

All of these constraints are handled as hard constraints within the convex optimization framework and are enforced directly by the CVXPYLayers solver. The mode variable handling the constraint on simultaneous battery charging and discharging is a convex relaxation, where the binary mode variable is relaxed to a continuous variable, preserving convexity and enabling differentiable optimization whilst at the same time avoiding too much cycling of the battery. No penalty-based or heuristic constraint handling techniques are employed.

\subsection{Decision-focused Learning}
\label{DFL}

The goal of this study is to create a forecasting model that helps to minimize our downstream objective function, Eq. (\ref{obj_func}). In order to do so, we create a neural network and attach an additional layer to it which solves the above optimization problem online, during the learning process. 

As mentioned, the neural network has to forecast the next 24 hours of PV generation, starting at midnight. The forecasting model used in our experiments is the Long Short-term Memory (LSTM) model. The reader is referred to \cite{ref33} for a detailed explanation on the inner workings of the LSTM model. The core principle behind the model is the use of two states, the cell and hidden state. These two states are updated on the basis of the states at the previous time step and the input associated with the current time step. These gates manage what gets passed on to the next states (and what forms the output).




The LSTM model forms the basis for both our regular forecaster and our DFL forecaster. We keep the hyper-parameters of both models exact to isolate the effect of the decision-focused loss versus a standard loss function. The hyperparameters of the LSTM model are taken from previous work using LSTM models to forecast residential PV\cite{ref6} in which extensive hyperparameter tuning was performed. These hyper-parameters were validated on our dataset by rerunning the hyperparameter tuning process using a limited hyper-parameter sample space and gave similar results, apart from the number of neurons (400 in \cite{ref6} versus 200 in this study). The optimal hyper-parameters and the sample values are shown in table \ref{tab:hyper} below.

\begin{table}
\begin{center}
\caption{Hyper-parameters for the LSTM}
\label{tab:hyper}
\begin{tabular}{| l | c | c | c |}
\hline
Hyper-parameter & Samples & Final value \\
\hline
Layers & [2, 3, 4] & 3 \\
Neurons & [200,300,400] & 200 \\
Dropout & [20\%, 30\%, 40\%, 50\%] & 50\% \\
Learning-rate & [1e-3, 1e-4, 1e-5] & 1e-4 \\
Batch-size & [32, 64, 128] & 32 \\
\hline
\end{tabular}
\end{center}
\end{table}

The CVXPY layer following the standard LSTM in the DFL model does not require any hyperparameters other than the optimization problem it represents. This layer does not require gradients and is only used for the sake of executing our convex optimization problem in a differentiable manner. The LSTM output, which is the PV generation for the next day, forms the input of this layer and the outputs are all the variables required to execute the optimization: the imported and exported energy, the battery energy, the value of the mode variable, and the charge and discharge values. 

Training this neural network works as depicted in Figure \ref{fig4}. For the baseline LSTM we remove the 'Optimize' step and go straight to Evaluate after getting the predictions. The evaluation is done using the Mean Squared Error (MSE), as shown in Eq. \ref{mse}. 

\begin{equation}
\begin{aligned}
\text{MSE} &= \frac{1}{n} \sum_{i=1}^{n} (y_i - \hat{y}_i)^2 \\
\label{mse}
\end{aligned}
\end{equation}

The MSE is a statistical metric that measures the quadratic difference between predictions and actual values. Therefore, if we use this metric to train our neural network, it will try to reduce the quadratic difference. Squaring the errors means that large outliers will be punished disproportionally, making the training process sensitive to trying to minimize these errors. Especially on rainy days this can make a difference, as PV output will be very peaked instead of smooth. In trying to avoid these errors, the neural network will tend to smooth / average out predictions and give a dampened and smoothened PV curve. 

Training the DFL forecaster works exactly as shown in Figure \ref{fig4}. There is more complexity behind the 'Optimize' step than what can be seen from the box in this figure however. The loss, which we call the regret, is the difference in the solution of the downstream optimization given the correct parameter values versus the parameter values given by the forecast model.

\begin{equation}
\begin{aligned}
\text{Regret}(\hat{y}, y) = f_{\hat{y}}(\mathcal{O}(\hat{y})) - f_{y}(\mathcal{O}(y)) \\
\label{regret}
\end{aligned}
\end{equation}

In Eq. \ref{regret}, $\mathcal{O}(\hat{y})$ is the optimal decision made for the optimization problem under parameters $\hat{y}$ and $f_{\hat{y}}(\mathcal{O}(\hat{y}))$ is the outcome of the downstream objective function given the optimal parameters. The regret is the difference between using the real values $y$ and the forecast values $\hat{y}$ in our problem. However, this requires an additional step that might not be apparent at first.  

First, we do a forward pass through our DFL forecasting model. The inputs given to the model are:

$
\mathbf{P_{pv}}^{\text{hist}}, 
\mathbf{DNI}^{\text{fcst}}, 
\mathbf{H_{cos}}^{\text{act}}, 
\mathbf{H_{sin}}^{\text{act}}, 
\mathbf{L}^{\text{fcst}}, 
\mathbf{Pr_{ex}}^{\text{act}}, 
\mathbf{Pr_{im}}^{\text{act}}
$. 

Where 'hist' indicates the previous 24 hours, 'fcst' a forecast for the next 24 hours, and 'act' known actuals for the next 24 hours. Prices, for example, are known up front because dynamic contract prices are typically made available in the afternoon the day before. The first four variables are used to forecast PV, which then gets fed into the CVX layer together with the final 3 variables to optimize the PV-B system. This gives us, among other outputs, the charge and discharge schedules, $\mathbf{P_{ch}}^{\text{fcst}}$ and $\mathbf{P_{dis}}^{\text{fcst}}$. Next, the optimization has to be executed again, with these charge and discharge schedules that we obtained and the actual PV data. So the inputs for this optimization problem are:

$
\mathbf{P_{pv}}^{\text{act}}, 
\mathbf{L}^{\text{fcst}}, 
\mathbf{Pr_{ex}}^{\text{act}}, 
\mathbf{Pr_{im}}^{\text{act}},
\mathbf{P_{ch}}^{\text{fcst}},
\mathbf{P_{dis}}^{\text{fcst}}
$. 

Solving this problem gives us $\mathcal{O}(\hat{y})$: the optimal cost given battery schedules obtained using $\hat{y}$. Finally, we execute the original optimization problem again using the correct PV actuals. The inputs are now:

$
\mathbf{P_{pv}}^{\text{act}}, 
\mathbf{L}^{\text{fcst}}, 
\mathbf{Pr_{ex}}^{\text{act}}, 
\mathbf{Pr_{im}}^{\text{act}}
$. 

This is basically the same optimization problem as in the first step, but now we do not need the inputs to predict PV and take the actual PV instead. Solving this problem gives us $\mathcal{O}(y)$, and we can now obtain the regret.

As a final remark, it is important to note that using battery schedules obtained with $\hat{y}$ can cause infeasible battery states when we plug them into an optimization problem using $y$. To avoid this, we relax the upper and lower limit of the battery SoC to 10\% and 90\% in Eq. \ref{SOC}, this way the optimization problem never became infeasible. 

\subsection{Experiments}

In our main experiment, we evaluate the performance of four different models: i) a naive seasonal model, predicting the next 24 hours to be the same as the last 24, ii) a standard two-phase LSTM model, iii) a cold-start DFL model in which we only use the regret to train the model (we call this 'DFL' in the results), and iv) a warm-start DFL model in which we use the state dictionary of the standard LSTM model as starting point and train on the regret after (which we call DFL-WS). 

The results of our experiments are evaluated both in terms of S-RMSE, the cost over the forecast period, and the relative cost. The S-RMSE is the RMSE of the PV forecast divided by the maximum PV in the forecast period and multiplied by 100, for each building, so as to avoid the impact of the size of the PV on this metric. The relative cost starts with a 0\% for the cost given perfect predictions and 100\% when we do not optimize (i.e. no battery is present and energy is bought when the net energy demand is positive and sold when the net energy demand is negative). The costs we obtain for our models fall in between these two and are extrapolated to percentages between no optimization and the perfect forecast. 

We can further substantiate the findings by performing a Diebold-Mariano test \cite{ref35}. The null hypothesis of this test poses that the forecast errors coming from the two forecast models bring about the same loss against the two-sided alternative that one of the models has a loss that is statistically higher or lower. Normally the test is performed to evaluate the difference in errors between forecast models, but it can also be used to evaluate the significance of the difference in costs. The test is described in more detail in Appendix A. We perform the test both across the buildings and on a per-building level for LSTM relative to DFL and relative to DFL-WS. 

It is important to mention that up to now we have not discussed how we handle the load forecast in our experiments. As our focus is on PV forecasting, we take a similar approach to how the DNI forecasts are handled, by adding an appropriate error to the load forecast. We use the same noise model as before with an $\alpha$ of minus 0.5 and a smoothing factor of 0.5. The $\beta$ in Eq. \ref{eq:mult_noise} is 50\% in our baseline experiment and the timeshift probability is 5\%. The baseline error that we artificially add to our load forecasts has a MAPE of c. 36\%, which lies between state-of-the-art research on residential load forecasting which proposes forecasts with MAPE of approximately 15-20\% (see for example \cite{ref34}), and a naive seasonal model, which averages a MAPE of approximately 75\% for our data. As an additional analysis to our main experiment, we further decrease and increase the load error to assess the impact on our optimal control by changing the $\beta$ in Eq. \ref{eq:mult_noise}.

\section{Results}

In this section we elaborate on the results from the four different models on the twenty Dutch households over the test period of approximately one and a half years. First, we go over the results from the base experimental set-up. After that, we look at the impact of a deterioration of the load forecast. Finally, we finish this section with a discussion of the results.

\subsection{Base experiments}
\label{base}

Table \ref{tab:results} summarizes our findings across the twenty buildings. In this table we present the average scaled RMSE (S-RMSE), the cost over the forecast period, and the relative cost. We can see that the LSTM model gives us the lowest S-RMSE, which makes sense, given that this model was trained on the MSE, so it directly tries to lower this metric. The decision-focused models, on the other hand, perform significantly worse in terms of RMSE. The DFL model has an average S-RMSE more than twice as high as LSTM. But DFL-WS model does mitigate this, improving the RMSE to the level of the naive model. 

\begin{table}
\begin{center}
\caption{Results averaged across the buildings}
\label{tab:results}
\begin{tabular}{| l | c | c | c |}
\hline
Model & S-RMSE & Real cost&Relative cost\\
 & &(in EUR)&(in \%)\\
\hline
No opt & n.a. & 499.8 & 100 \\
Naive & 13.8 & 393.1 & 51.1 \\
LSTM & \textbf{8.2} & 376.6 & 43.6 \\
DFL & 19.9 & 368.7 & 40.0 \\
DFL-WS & 13.7 & \textbf{358.7} & \textbf{35.4} \\
Perfect & 0.0 & 281.4 & 0.0 \\ 
\hline
\end{tabular}
\end{center}
\end{table}

But what does this mean in terms of cost? Having no battery installed (no optimization) gives us an average cost of about 500 EUR across the test period and performing optimal control with perfect knowledge of PV gives an average cost of 281 EUR. The best performing model in terms of cost is the DFL-WS model which, at 359 EUR, gets about two-thirds along the way between perfect knowledge and no optimization. The two-phase LSTM model on the other hand, which performed best in terms of S-RMSE, performs about 8\% worse in terms of cost, giving us a total average cost of 377 EUR. 

\begin{figure}[!t]
\centering
\includegraphics[width=\textwidth]{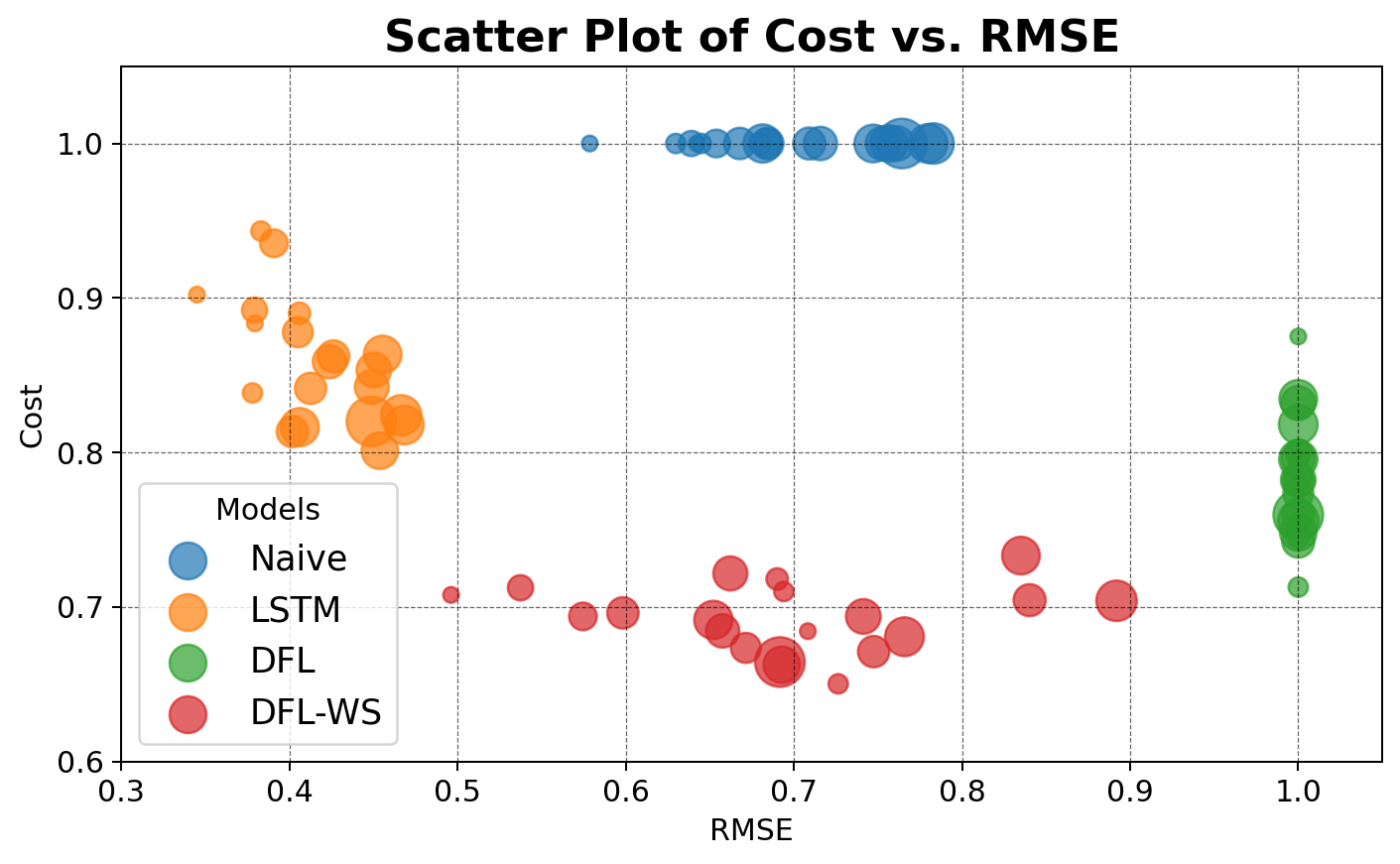}
\caption{A scatter plot of the results plotting the RMSE and cost relative to the perfect forecast (which we set to 0) and the worst performing model which is set as 1. Each dot represents a building and the size of the dot represents the size of the battery.}
\label{fig5}
\end{figure}

The significance test results are shown in tables \ref{tab:dm_summary} and \ref{tab:dm_summary_error} for cost and accuracy respectively. Both the DFL and DFL-WS method have significantly lower costs than the LSTM model (the negative DM value indicates that LSTM costs are higher). However, for the DFL method at a per building level, while always negative, only 7 out of 20 buildings have statistically lower costs. When warm-starting is performed the p-value is statistically significant at a per building level for every single building (at p < 0.001 for 18 out of 20 buildings and at p < 0.01 for the remaining 2). With regards to the error the reverse is true. The LSTM models are significantly lower in error, both overall as well as across all buildings. This is visually also very clear in figure \ref{fig5} as all the yellow dots lie to the left of the red and green dots.

\begin{table}
\begin{center}
\caption{Diebold-Mariano test for statistical significance of difference in cost}
\label{tab:dm_summary}
\begin{tabular}{| l | c | c | c |}
\hline
\textbf{Comparison} & \textbf{DM value} & \textbf{$p$-value} & \textbf{Significant buildings*} \\
\hline
LSTM vs. DFL & -6.13 & < 0.001** & 7/20 \\
LSTM vs. DFL-WS & -20.17 & < 0.001** & 20/20 \\
\hline
\addlinespace
\multicolumn{4}{l}{\footnotesize * Number of individual buildings with statistically significant cost savings ($p < 0.05$)} \\
\multicolumn{4}{l}{\footnotesize ** Significant at the 0.001 level.} \\
\end{tabular}
\end{center}
\end{table}

\begin{table}
\begin{center}
\caption{Diebold-Mariano test for statistical significance of difference in error}
\label{tab:dm_summary_error}
\begin{tabular}{| l | c | c | c |}
\hline
\textbf{Comparison} & \textbf{DM value} & \textbf{$p$-value} & \textbf{Significant buildings*} \\
\hline
LSTM vs. DFL & 176.84 & < 0.001** & 20/20 \\
LSTM vs. DFL-WS & 133.58 & < 0.001** & 20/20 \\
\hline
\addlinespace
\multicolumn{4}{l}{\footnotesize * Number of individual buildings with statistically significant error increases ($p < 0.05$)} \\
\multicolumn{4}{l}{\footnotesize ** Significant at the 0.001 level.} \\
\end{tabular}
\end{center}
\end{table}

Figure \ref{fig5} plots each household as a dot on a relative scale for RMSE and cost with 1 being the worst performing model and 0 being the control with perfect knowledge. The size of the dot represents the size of the battery in the households. First, we can clearly see that the DFL-WS model outperforms the LSTM model for all buildings in terms of cost (just as the LSTM model outperforms all other models in terms of RMSE). Second, we can see that the spread in RMSE across the buildings is very large for the DFL-WS models but the spread in costs is not, while the LSTM models are rather tightly nit both in RMSE and cost. This gives us an indication that minimizing the RMSE does not necessarily coincide with minimizing the cost as the DFL-WS models started from the state dictionary of the LSTM models and where further trained from this starting point to minimize costs. Third, we can see that the battery size as an impact for the LSTM costs. A bigger battery means slightly lower cost, relative to the naive model. However, for the decision-focused models the impact is not as clear-cut. Finally, it is worth noting that the DFL model also outperforms the LSTM for all buildings, except one for which the cost remained almost the same.

\subsection{Value creation}

\begin{figure*}[!t]
\centering
\includegraphics[width=\textwidth]{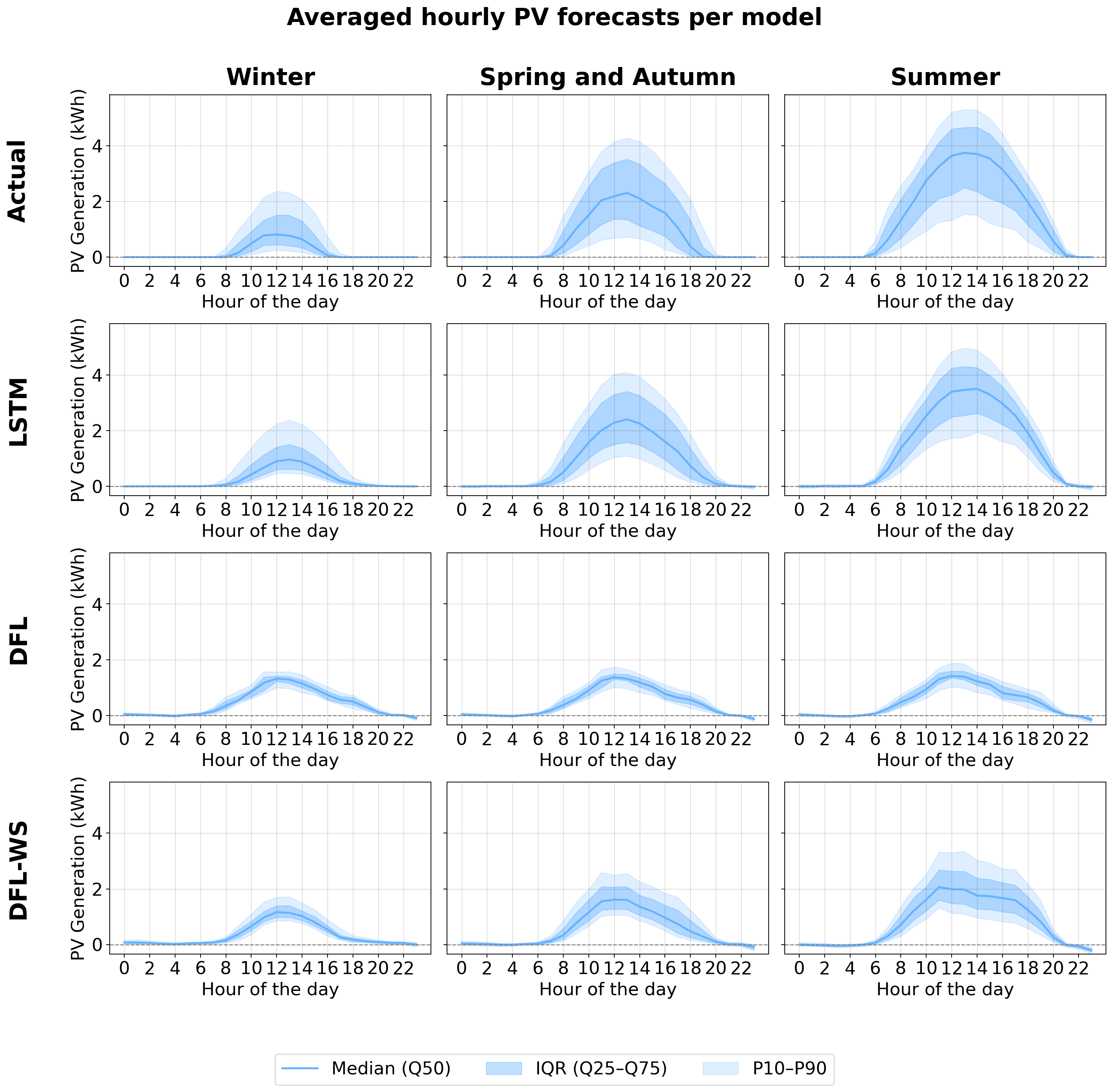}
\caption{PV power forecasts per hour in kWh, averaged across the buildings. The left column has data from the months November to February, the right column May to August and the middle column the remaining months. The shaded areas represent the inter quartile range (IQR) and the range of the 10th percentile until the 90th percentile of forecasts.}
\label{fig6}
\end{figure*}

Figure \ref{fig6} plots the PV forecast across the households on an averaged hourly basis, per model. The shaded areas represent the IQR and P10 to P90. From this figure, we can see that the forecasts from the LSTM model look very similar to the actual data. The DFL forecasts, on the other hand, look very different. During winter, the median forecast is slightly higher than the actuals, and during the other seasons the forecasts are significantly lower. The spread in forecasts is also much smaller for the DFL models, indicating a preference for consistent forecasts. 

\begin{figure*}[!t]
\centering
\includegraphics[width=\textwidth]{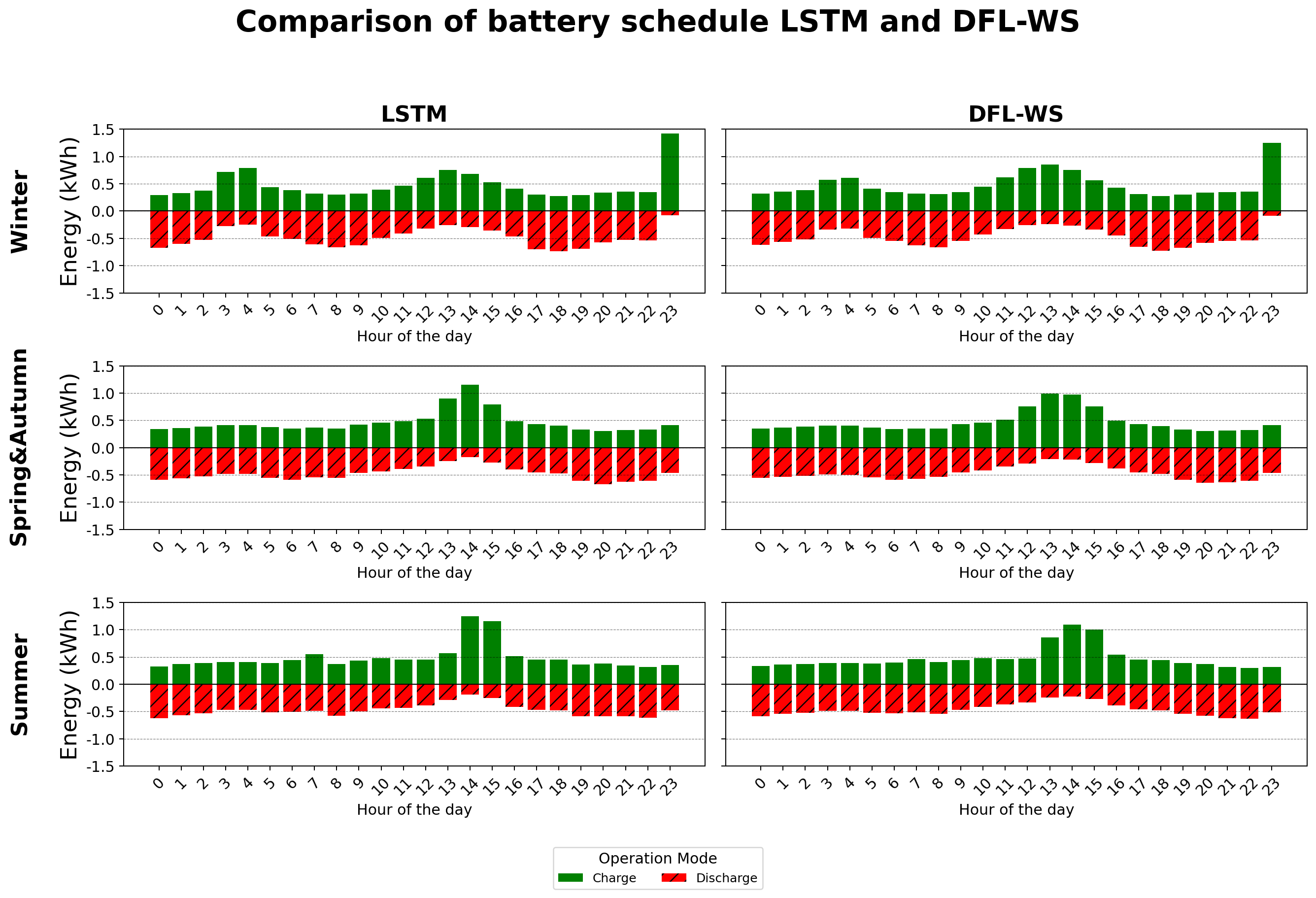}
\caption{Charge (positive) and discharge (negative) schedules per hour, averaged across the buildings. The months making up the seasonal plots are the same as the PV power forecasts. The large charging peak in the winter clearly indicates that the models are better adapted to the other seasons.}
\label{fig7}
\end{figure*}

The way this translates into battery behavior can be seen in Figure \ref{fig7}. We can see that in winter the battery will be charged more during the day using DFL-WS (right) while the LSTM model (left) promotes charging during the early hours when prices are low. Because of this increased charging during the day, less will be needed in the end to get the SoC back to 50\% as well. During the other months, we can see that this pattern changes, as the LSTM model predicts higher average PV. The DFL model charges more between 10 and 13 o' clock while the LSTM model charges a bit later, at 14 and 15 o' clock. In the summer period we can again also see more charging during the morning, between 6 and 7, which leads to a bit more discharging between 10 and 13 o' clock. It is worth noting that most value comes from the summer period, where DFL-WS performs 11\% better than LSTM, while it performs 5\% better in the spring and autumn period and only 1\% better in winter months.

It is interesting to note that the main increase in performance of our battery comes in the summer period, where the DFL-WS model performs 11\% better, compared to the winter when there was almost no out-performance. In the summer months there is a lot more PV in general, so the potential cost savings are also higher, further indicating that the DFL model indeed aligns well with the downstream task. 

\subsection{Impact of load forecast quality}
\label{load}

Finally, the role that the load forecast plays on the DFL forecast is assessed. In our PV-B system, both PV and load are uncertain parameters. Up until now, we have assumed a load forecast of a certain quality by adding a noise factor to the actual load and using it as such. But the quality of the PV forecast is dependent on the quality of our load forecast in the DFL scenario, so how does this impact our results relative to a two-phased approach? 

\begin{figure}[!t]
\centering
\includegraphics[width=\textwidth]{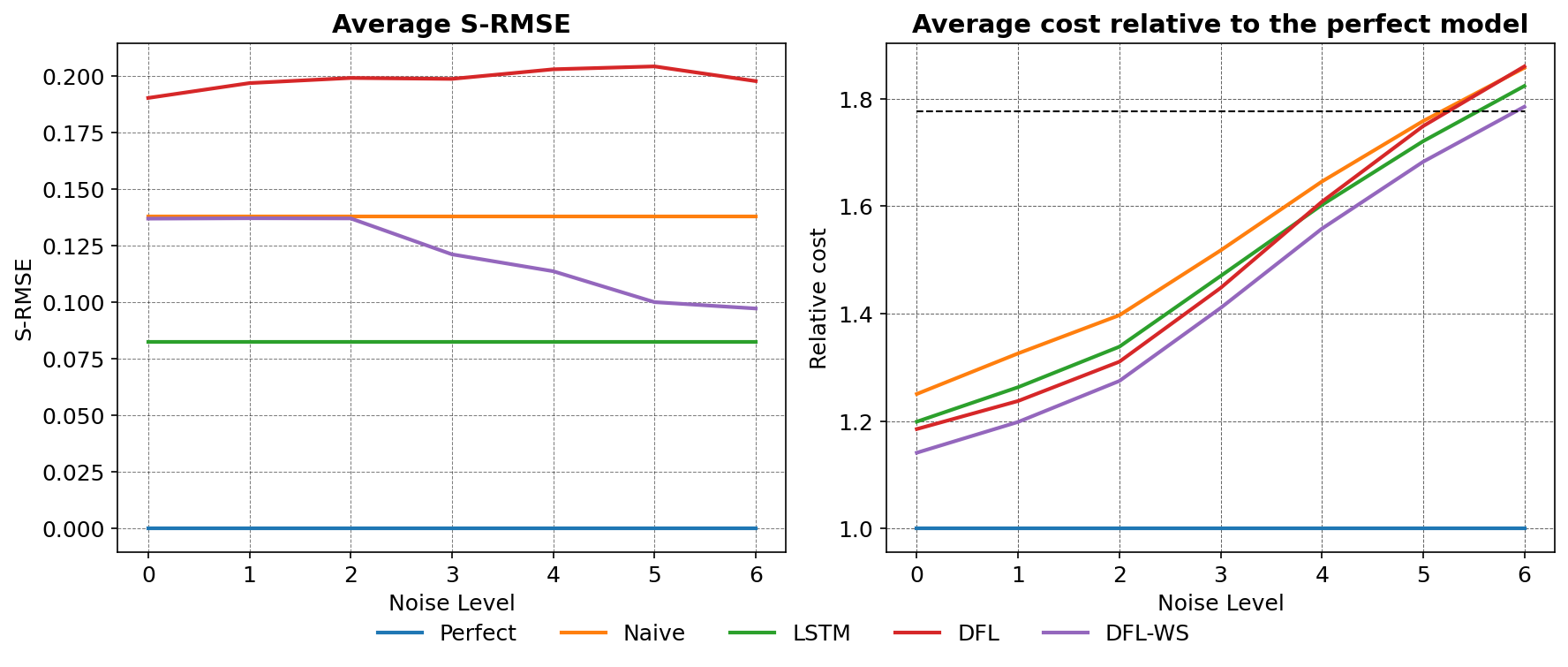}
\caption{The impact of the load error on the MSE and downstream cost, given the use of a specific model. Both plots are scaled, for the cost this means that the perfect forecast has a cost set to 1. A cost of 1.2 in this case means 20\% worse than the perfect forecast. The dashed line indicates the cost of not optimizing.}
\label{fig8}
\end{figure}

Figure \ref{fig8} plots how the average RMSE and cost increase when we add increasing amounts of noise to the load forecast. As an indication of what these noise levels represent: 0 noise means a perfect forecast. Level 2 is what we used in the baseline analyzes earlier (these numbers align with Table \ref{tab:results}). In this case, the RMSE of the load forecast is 0.32 and the MAPE is 36.11\%. At level 4 the RMSE is 0.59 and the MAPE 79.81\% and at level 6 it is 0.8 and 128.12\%. 


These plots provide us with some interesting findings. In terms of the RMSE, the impact is limited to the DFL models, as two-phase approaches do not take the load into account when forecasting PV. When we look at the DFL model, we can see that (significantly) increasing the error of the load forecast has a muted effect on the RMSE of the DFL model, increasing it slightly. Interestingly, for the DFL-WS model it has the opposite effect. This is because the DFL-WS model starts from the situation of the LSTM model, and the fine-tuning in a DFL manner is far more limited when the load forecast deteriorates significantly. The average number of epochs the DFL-WS model trains with noise level 6 is 20 before reaching the best regret on the test set, while it is 52 when the noise level is 2 (the baseline from before). 

What this means in terms of cost can be seen in the bottom plot. The cost increases similarly for the naive, LSTM and DFL-WS models, so relatively the gain in value of using a DFL-WS vanishes with high errors. However, it is worth noting that the errors are very high, a lot higher than state-of-the-art models, when we enter noise levels of 4 and above. Recall from table \ref{tab:results} that 'no optimization' gave a cost of approximately 500 EUR over the test period. This is indicated by the dashed line in the cost plot, so at noise level 6 it is not even worth optimizing at all. Furthermore, it is interesting to note that the DFL model loses more in performance relative to the other models. While the DFL-WS model is always best, at high noise levels the DFL model performs worse than LSTM, and even worse than a naive model. Given that the load forecast is a parameter that is directly used during training, it would make sense that it has a more prominent effect on quality deterioration than for a two-phase model, but importantly these results show us that warm-starting a DFL model mutes the impact to quality limitations in other stochastic inputs and even though performance deteriorates, the model using DFL still outperforms a two-phase approach when we warm-start.    

It is worth noting that, while the DFL model does become the worst performing model at our noise level of six, this noise level comes with a MAPE of 128.12\% which is even worse than a naive seasonal model applied to the load (which falls somewhere between noise levels 3 and 4). If we focus on the noise levels 1 to 3, where you would expect most modern load forecast models to land in, the DFL models always outperform the two-phase models.

\section{Discussion}
Our results demonstrate that an integrated approach to forecasting and optimization aligns the forecasting model with the downstream task, leading to an out-performance of approximately 8\% over traditionally trained forecasting models, in the specific use-case of a PV-B system. This confirms our primary contribution, that forecasting models trained on task-specific metrics outperform generic forecasting models in terms of costs, albeit at the cost of general applicability.

Next, we showed that warm-starting these models proves to be beneficial and further improves the cost advantage. We can see this as a way of helping the model by providing it an indicative/typical shape of the target variable after which the model is then tasked to improve upon it by aiming to decrease downstream costs. The RMSE once again increases after fine-tuning, but it is better than a cold-start DFL model, which improves its statistical robustness. Warm-starting also helps with the generalization possibilities of the forecasts. One of the main disadvantages of DFL models is that they are uniquely trained to characteristics of the optimisation problem. The size of the battery, for example, is a mandatory parameter for training but is specific to the household at hand, therefore each DFL model can only be used on that specific household. Warm starting makes it so a general PV forecasting model for a specific region can be trained and then used as a template for all households in that region for further finetuning using DFL. 

We also evaluated the impact of other stochastic parameters on our model performance. The household load is also unknown in advance, and sometimes energy prices can also be unknown, for example if we look further ahead and make a weekly schedule. It makes sense that the quality of these forecasts also impact the quality of our PV forecast when we apply a DFL framework, as the load is a parameter that is used to assess the loss (our regret function) of our PV model. Interestingly, we find that DFL models are not impacted more harshly by a deterioration in quality of the load forecast when a warm-start methodology is applied. 

\begin{figure}[!t]
\centering
\includegraphics[width=\textwidth]{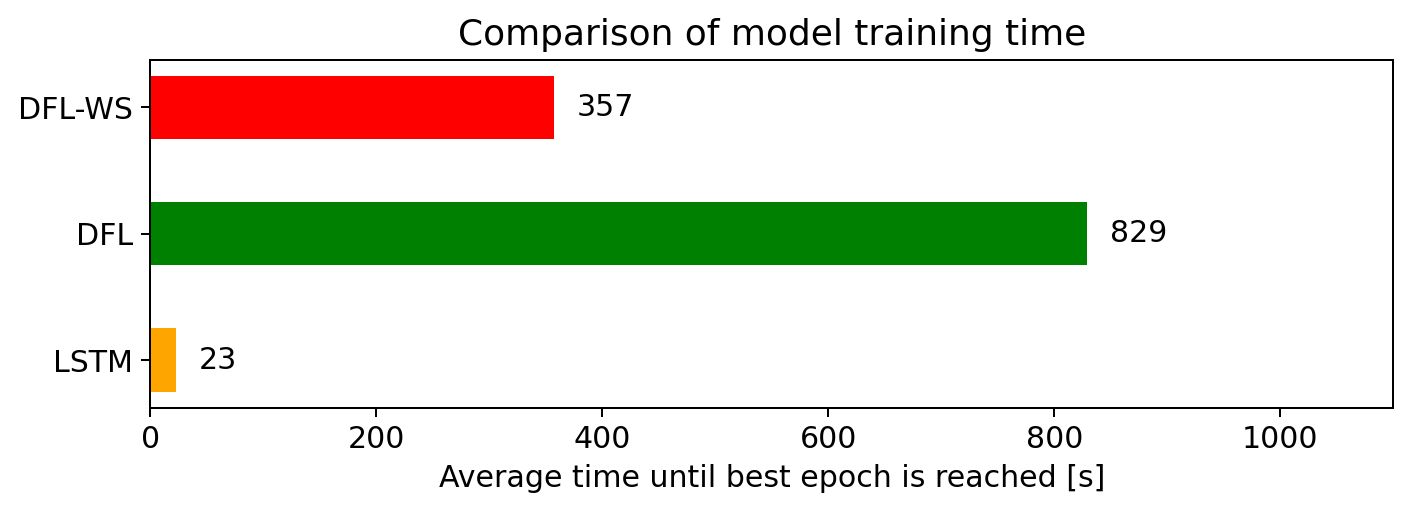}
\caption{The average training time (in seconds) until the best performing epoch (based on test set performance) is reached for each of the three models.}
\label{fig9}
\end{figure}

Going beyond our objectives, it is important to mention the computational impact of including an optimization problem in the training process of a neural network. In relative terms, the DFL models require a lot more computational power to train compared to two-phase models\footnote{The computer these models are trained on has an AMD Ryzen 7 7800X3D 8-Core Processor, 32GB of RAM and an NVIDIA GeForce RTX 4070 SUPER GPU.}. On average, it took 36 times longer to train the DFL models and 16 times longer to finetune the DFL-WS models compared to training the baseline LSTM. In absolute terms, this is all still easily manageable given that our optimization problem is fairly simple; the LSTM model training typically took 23 seconds while it took 5 minutes and 57 seconds to further finetune the DFL-WS model (see Figure \ref{fig9}). The lower computational cost (2.3 times faster) from only finetuning the models (DFL-WS) instead of training from scratch (DFL) is an additional advantage of warm starting, on top of the increased performance. This decrease in computational cost comes from the faster convergence of these models, typically only requiring a fraction of the number of epochs to train. Finally, It is also worth noting that a lot depends on the solver included in the DFL neural network training. Although we have not investigated this for our research, Wahdany et al.\cite{ref25} found that the solver used by the free CVXPY package performs worse than commercial alternatives such as GUROBI, by a factor of more than 50. 

Our findings suggest that practitioners should transition from purely statistical forecasting objectives to decision-focused learning frameworks when managing home energy management systems (HEMS). By employing a warm-start strategy, managers can reconcile the need for general statistical reliability with the economic requirement of task-specific optimization. This methodology utilizes a robust statistical forecast as a baseline template, which is then refined to account for the household specific cost structures. Furthermore, this approach offers significant operational scalability compared to cold-start DFL methods: by using a pre-trained model as a universal starting point, the training process is reduced to fine-tuning. This allows for the deployment of custom models across diverse households without the computational overhead of full-scale retraining. Of course, the computational cost is still higher than (re-)training a standard neural network. We discuss this in the next section.

\section{Limitations and future research directions}

We showed that the DFL method comes with an S-RMSE of c. 20\%, which could raise concerns regarding potential operational risks. In the methodology section, we mentioned relaxing our SoC limits from 20\% and 80\% to 10\% and 90\%. In doing so, the constraints were never violated in any of the test sets. So, at a local level the deviations never breached any physical limits. However, on a broader grid level it could be argued that these methods, prioritizing cost over accuracy, could lead to less predictable prosumer behavior and more uncertainty for the regulator with regard to injection and off-take patterns. This balancing exercise between personal gains and system stability is an important trade-off to consider, which we have not looked at in this study. Additionally, this study was limited to 20 Dutch households in close proximity to each other. This raises questions about the generalization of the results to other climate zones. 

Beyond these limitations, several promising avenues for future research remain. Future research can look at the impact of the complexity of the optimization problem on the forecast results (and the computational cost). See, for example, the work of Despeghel et al.\cite{ref36} for a convex model taking into account battery losses. Another avenue of future research can look at the use of different downstream objective functions, for example, the cost of emissions, or balancing when looking from the perspective of a grid operator. These are just a couple of possible research directions, but plenty more are possible in this young field. 

\section{Conclusions}
Our research can guide practitioners towards a more informed decision process when scheduling their battery in a PV-B system. In this study, we focused on forecasting residential PV. We apply the forecasts to a PV-B system, focused on driving down costs. The goal was to use an integrated approach in which the downstream task of scheduling a PV-B system is taken into account when training our forecasting model. 

Across 20 residential buildings, this integrated approach yielded consistent cost improvements over standard forecasting models. Cold-start decision-focused models provided a modest cost advantage of 3.6\%. By contrast, warm-starting the decision-focused model using a statistically trained forecast substantially improved performance. This approach increased the average cost advantage to 7.8\%, resulted in superior performance for all 20 buildings, and mitigated the degradation in statistical accuracy.

Importantly, the warm-start strategy also strengthened the statistical robustness of the results: performance gains were statistically significant both across buildings and at the individual building level, addressing a key practical concern associated with cold-start decision-focused learning. Taken together, these findings demonstrate that warm-started decision-focused learning can reconcile the trade-off between statistical reliability and economic optimality in residential energy management. Additionally, warm-starting mitigates the computational overhead relative to cold-starting, a key weakness of DFL methods.

More broadly, this work highlights the value of evaluating forecasting models in the context of their intended downstream operational tasks. For PV-B systems, the combination of strong statistical forecasts with specific decision-aware fine-tuning offers a scalable and economically meaningful option for more effective battery operation. This framework extends beyond PV forecasting and is applicable to a wide range of energy system optimization problems where forecasts directly impact operational decisions.

\section*{Acknowledgments}
Hussain Kazmi acknowledges support from Research Foundation – Flanders (FWO), Belgium (research fellowship 1262921N).



\appendix

\section*{Appendix A: Diebold--Mariano Test}
\label{appendix1}

In this appendix we describe the Diebold--Mariano (DM) test used to compare
the predictive performance of two forecasting methods in terms of downstream
costs.

Let $C_{A,t}$ and $C_{B,t}$ denote the cost incurred by forecasting
methods $A$ and $B$ at time $t$, respectively. The loss differential is defined
as
\begin{equation}
d_t = C_{A,t} - C_{B,t}.
\end{equation}

The null hypothesis of equal predictive performance is
\begin{equation}
H_0 : \mathbb{E}[d_t] = 0,
\end{equation}
against the two-sided alternative hypothesis that method $A$ yields lower costs,
$\mathbb{E}[d_t] < 0$, or higher costs $\mathbb{E}[d_t] > 0$.

Given $T$ forecast evaluation periods, the sample mean of the loss differential
is
\begin{equation}
\bar{d} = \frac{1}{T} \sum_{t=1}^{T} d_t.
\end{equation}

Because the loss differential may exhibit serial correlation, the DM test
employs a heteroskedasticity- and autocorrelation-consistent(HAC) estimator of the
long-run variance, as proposed in \cite{ref37},
\begin{equation}
\hat{\sigma}^2_d = \gamma_0 + 2 \sum_{k=1}^{h-1} \gamma_k,
\end{equation}
where
\begin{equation}
\gamma_k = \frac{1}{T} \sum_{t=k+1}^{T}
(d_t - \bar{d})(d_{t-k} - \bar{d}),
\end{equation}
and $h$ denotes the number of overlapping forecast periods (h=1 in our case).

The Diebold--Mariano test statistic is given by
\begin{equation}
DM = \frac{\bar{d}}{\sqrt{\hat{\sigma}^2_d / T}},
\end{equation}
which is asymptotically standard normally distributed under the null
hypothesis ($DM \sim \mathcal{N}(0, 1)$), meaning the further the values are away from 0, the lower the p-value and the stronger the evidence against the null hypothesis.


\begin{thebibliography}{00}

\bibitem{ref1}
European Union, “Directive (EU) 2023/2413 of the European Parliament and the Council,” *Official Journal of the European Union*, pp. 1–77, Oct. 2023. [Online]
. Available: https://eur-lex.europa.eu/eli/dir/2023/2413/oj

\bibitem{ref2}
K. Bódis, I. Kougias, A. Jäger-Waldau, N. Taylor, and S. Szabó, “A high-resolution geospatial assessment of the rooftop solar photovoltaic potential in the European Union,” *Renewable and Sustainable Energy Reviews*, vol. 114, pp. 1–13, Oct. 2019. DOI: https://doi.org/10.1016/j.rser.2019.109309.

\bibitem{ref3}
C. Scott, M. Ahsan, and A. Albarbar, “Machine learning for forecasting a photovoltaic (PV) generation system,” *Energy*, vol. 278, pp. 127807, 2023. DOI: https://doi.org/10.1016/j.energy.2023.127807.

\bibitem{ref4}
M. Kaffash, K. Bruninx, and G. Deconinck, “Data-driven forecasting of local PV generation for stochastic PV-battery system management,” *Int. J. Energy Res.*, vol. 45, no. 11, pp. 15962–15979, 2021. DOI: https://doi.org/10.1002/er.6826.

\bibitem{ref5}
I. Tavares et al., “Comparison of PV power generation forecasting in a residential building using ANN and DNN,” *IFAC-PapersOnLine*, vol. 55, no. 9, pp. 291–296, 2022.

\bibitem{ref6}
J. Depoortere, J. Driesen, J. Suykens, and H. S. Kazmi, “SolNet: Open-source deep learning models for photovoltaic power forecasting across the globe,” *International Journal of Forecasting*, vol. 41, no. 3, pp. 1223–1236, 2025. DOI: https://doi.org/10.1016/j.ijforecast.2024.12.003.

\bibitem{ref7}
J. Bottieau et al., “A cross-learning approach for cold-start forecasting of residential photovoltaic generation,” *Electric Power Systems Research*, vol. 212, pp. 108415, 2022. DOI: https://doi.org/10.1016/j.epsr.2022.108415.

\bibitem{ref8}
M. Perera, J. De Hoog, K. Bandara, and S. Halgamuge, “Multi-resolution, multi-horizon distributed solar PV power forecasting with forecast combinations,” *Expert Systems with Applications*, vol. 205, pp. 117690, 2022. DOI: https://doi.org/10.1016/j.eswa.2022.117690.

\bibitem{ref9}
H. Kazmi, C. Fu, and C. Miller, “Ten questions concerning data-driven modelling and forecasting of operational energy demand at building and urban scale,” *Building and Environment*, vol. 239, pp. 110407, 2023. DOI: https://doi.org/10.1016/j.buildenv.2023.110407.

\bibitem{ref10}
K. Barhmi, C. Heynen, S. Golroodbari, and W. van Sark, “A review of solar forecasting techniques and the role of artificial intelligence,” *Solar*, vol. 4, no. 1, pp. 99–135, Feb. 2024.

\bibitem{ref11}
S. Lim, J. Huh, S. Hong, C. Park, and J. Kim, “Solar Power Forecasting Using CNN-LSTM Hybrid Model,” *Energies*, vol. 15, no. 21, pp. 8233, 2022. DOI: https://doi.org/10.3390/en15218233.

\bibitem{ref12}
D. Yang et al., “Verification of deterministic solar forecasts,” *Solar Energy*, vol. 210, pp. 20–37, 2020.

\bibitem{ref13}
A. Nottrott, J. Kleissl, and B. Washom, “Energy dispatch schedule optimization and cost benefit analysis for grid-connected, photovoltaic-battery storage systems,” *Renewable Energy*, vol. 55, pp. 42–52, Jul. 2013. DOI: https://doi.org/10.1016/j.renene.2012.12.036.

\bibitem{ref14}
B. Zou, J. Peng, R. Yin, Z. Luo, J. Song, T. Ma, S. Li, and H. Yang, “Energy management of the grid-connected residential photovoltaic-battery system using model predictive control coupled with dynamic programming,” *Energy and Buildings*, vol. 279, pp. 1–22, Dec. 2023. DOI: https://doi.org/10.1016/j.enbuild.2022.112712.

\bibitem{ref15}
S. Zhang and Y. Tang, “Optimal schedule of grid-connected residential PV generation systems with battery storages under time-of-use and step tariffs,” *Journal of Energy Storage*, vol. 23, pp. 1–8, Jun. 2019. DOI: https://doi.org/10.1016/j.est.2019.01.030.

\bibitem{ref16}
Z. Song, X. Guan, and M. Cheng, “Multi-objective optimization strategy for home energy management system including PV and battery energy storage,” *Energy Reports*, vol. 8, pp. 6971–6986, Nov. 2022. DOI: https://doi.org/10.1016/j.egyr.2022.04.023.

\bibitem{ref17}
Z. Luo, J. Peng, Y. Tan, R. Yin, B. Zou, M. Hu, and J. Yan, “A novel forecast-based operation strategy for residential PV-battery-flexible loads systems considering the flexibility of battery and loads,” *Energy Conversion and Management*, vol. 278, pp. 1–19, Feb. 2023. DOI: https://doi.org/10.1016/j.enconman.2023.116705.

\bibitem{ref18}
X. Liang, W. Ge, Z. Zhang, F. Zheng, X. Jin, and Z. Du, “Two-stage optimal scheduling for flexibility and resilience tradeoff of PV-battery building via smart grid communication,” *Sustainable Cities and Society*, vol. 116, pp. 105919, 2024.

\bibitem{ref19}
J. Kotary, F. Fioretto, P. Van Hentenryck, and B. Wilder, “End-to-end constrained optimization learning: A survey,” *arXiv preprint*, arXiv:2103.16378, 2021.

\bibitem{ref20}
A. Agrawal, B. Amos, S. Barratt, S. Boyd, S. Diamond, and Z. Kolter, “Differentiable convex optimization layers,” in *Advances in Neural Information Processing Systems*, vol. 32, pp. 9562–9581, 2019. 

\bibitem{ref21}
P. Van Hentenryck, “Optimization Learning,” *arXiv preprint*, arXiv:2501.03443, 2025.

\bibitem{ref22}
C. Bergmeir et al., “Predict+ Optimize Problem in Renewable Energy Scheduling,” *IEEE Access*, 2025.

\bibitem{ref23}
Á. Paredes, J. F. Toubeau, J. A. Aguado, and F. Vallée, “On the participation of energy storage systems in reserve markets using Decision Focused Learning,” *Sustainable Energy, Grids and Networks*, vol. 42, pp. 101677, 2025.

\bibitem{ref24}
J. Depoortere, H. S. Kazmi, and J. Driesen, “Using value-oriented forecasting to optimize PV battery system operation,” in *Proceedings of the 16th ACM International Conference on Future and Sustainable Energy Systems (E-Energy '25)*, pp. 936–940, 2025. DOI: https://doi.org/10.1145/3679240.3734666.

\bibitem{ref25}
D. Wahdany, C. Schmitt, and J. L. Cremer, “More than accuracy: end-to-end wind power forecasting that optimises the energy system,” *Electric Power Systems Research*, vol. 221, pp. 109384, 2023.

\bibitem{ref26}
C. Dupont, P. Favaro, F. Vallée, B. Francois, and J. F. Toubeau, “Decision-Focused Learning for Optimized Participation of an Industrial Consumer in Energy-Only and Reserve Markets,” in *2024 IEEE PES Innovative Smart Grid Technologies Europe (ISGT EUROPE)*, pp. 1–5, Oct. 2024.

\bibitem{ref27}
A. Stratigakos, A. Michiorri, and G. Kariniotakis, “A Value-Oriented Price Forecasting Approach to Optimize Trading of Renewable Generation,” in *2021 IEEE Madrid PowerTech*, Madrid, Spain, pp. 1–6, 2021. DOI: 10.1109/PowerTech46648.2021.9494832.

\bibitem{ref28}
J. Han, L. Yan, and Z. Li, “A Task-Based Day-Ahead Load Forecasting Model for Stochastic Economic Dispatch,” *IEEE Transactions on Power Systems*, vol. 36, no. 6, pp. 5294–5304, Nov. 2021. DOI: 10.1109/TPWRS.2021.3072904.

\bibitem{ref29}
Y. Zhang, M. Jia, H. Wen, and Y. Shi, “Toward Value‑Oriented Renewable Energy Forecasting: An Iterative Learning Approach,” *IEEE Transactions on Smart Grid*, 2025. DOI: 10.1109/TSG.<actual suffix>.10771620.

\bibitem{ref30}
J. Weniger, T. Tjaden, and V. Quaschning, “Sizing of Residential PV Battery Systems,” *Energy Procedia*, vol. 46, pp. 78–87, 2014. 8th International Renewable Energy Storage Conference and Exhibition (IRES 2013). DOI: https://doi.org/10.1016/j.egypro.2014.01.160.

\bibitem{ref31}
Z. T. Olivieri and K. McConky, “Optimization of residential battery energy storage system scheduling for cost and emissions reductions,” *Energy and Buildings*, vol. 210, pp. 109787, 2020.

\bibitem{ref32}
P. Zippenfenig, “Open-Meteo.com Weather API,” *Software*, 2023. DOI: https://doi.org/10.5281/zenodo.7970649. Available: https://open-meteo.com/

\bibitem{ref33}
S. Hochreiter and J. Schmidhuber, “Long short-term memory,” *Neural Computation*, vol. 9, no. 8, pp. 1735–1780, 1997.

\bibitem{ref34}
H. Zang, R. Xu, L. Cheng, T. Ding, L. Liu, Z. Wei, and G. Sun, “Residential load forecasting based on LSTM fusing self-attention mechanism with pooling,” *Energy*, vol. 229, pp. 120682, 2021.

\bibitem{ref35}
F. X. Diebold and R. S. Mariano, ``Comparing predictive accuracy,'' \textit{Journal of Business \& Economic Statistics}, vol. 13, no. 3, pp. 253--263, 1995.

\bibitem{ref36}
J. Despeghel, J. Tant, and J. Driesen, “Convex optimization of PV-battery system sizing and operation with non-linear loss models,” *Applied Energy*, vol. 353, pp. 121976, 2024.

\bibitem{ref37}
W. K. Newey and K. D. West, ``A simple, positive semi-definite, heteroskedasticity and autocorrelation consistent covariance matrix,'' \textit{Econometrica}, vol. 55, no. 3, pp. 703--708, 1987.



\end{thebibliography}
\end{document}